\pdfoutput=1
\documentclass[a4page,10pt]{article}
\usepackage[top=0.8in, bottom=0.8in, left=0.5in, right=0.5in]{geometry}
\usepackage{fancyhdr}
\renewcommand{\thispagestyle}[1]{}
\usepackage{times,amsmath,graphicx}
\usepackage[ruled, vlined, boxed]{algorithm2e}

\def\urltilda{\kern -.15em\lower .7ex\hbox{\~{}}\kern .04em}

\author{Ashwin Srinivasan $^1$, Gautam Shroff $^2$, Lovekesh Vig $^2$,  Sarmimala Saikia $^2$, Puneet Agarwal $^2$\\
     \small ashwin.srinivasan@wolfson.oxon.org, gautam.shroff, lovekesh.vig, sarmimala.saikia, puneet.a@tcs.com  \\
     \small $^1$Department of Computer Science \& Information Systems  \\
     \small BITS-Pilani, Goa Campus, Goa 403726.\\
     \small $^2$TCS Research, New Delhi. 
}

\date{}

\title{Generation of Near-Optimal Solutions Using ILP-Guided Sampling}

\begin{document}
\maketitle

\begin{abstract}

Our interest in this paper is in optimisation problems that are intractable
to solve by direct numerical optimisation, but nevertheless have significant
amounts of relevant domain-specific knowledge. The category of heuristic search
techniques known as estimation of distribution algorithms (EDAs) seek to
incrementally sample from probability distributions in which
optimal (or near-optimal) solutions have increasingly higher probabilities.
Can we use domain knowledge to assist the estimation of these distributions?
To answer this in the affirmative, we need: (a) a general-purpose technique for the
incorporation of domain knowledge when constructing models for optimal values; and (b)
a way of using these models to generate new data samples. Here we investigate
a combination of the use of Inductive Logic Programming (ILP) for (a), and standard logic-programming
machinery to generate new samples for (b). Specifically, on each iteration of distribution estimation, an
ILP engine is used to construct a model for good solutions. The resulting theory is then used to guide
the generation of new data instances, which are now restricted to those derivable using the ILP model in
conjunction with the background knowledge). We demonstrate the approach
on two optimisation problems (predicting optimal depth-of-win for the KRK endgame, and job-shop
scheduling). Our results are promising: (a) On each iteration of distribution estimation,
samples obtained with an ILP theory have a substantially
greater proportion of good solutions than samples without a theory; and (b) On
termination of distribution estimation, samples obtained with an ILP theory contain more near-optimal
samples than samples without a theory.
Taken together, these results suggest that the use of ILP-constructed theories
could be a useful technique for incorporating complex domain-knowledge into estimation distribution procedures.

\end{abstract}

\section{Introduction}
\label{sec:intro}

There are many real-world planning problems for which domain knowledge
is qualitative, and not easily encoded in a form suitable for
numerical optimisation. Here, for instance, are some
guiding principles that are followed by the Australian Rail Track Corporation when
scheduling trains:
(1)  If a ``healthy'' Train is running late, it should be given
equal preference to other healthy Trains;
(2) A higher priority train should be given preference to a lower
priority train, provided the delay to the lower priority train is
kept to a minimum; and so on. It is evident from this that train-scheduling
may benefit from knowing if a train is ``healthy'', what a train's priority is, and so on.
But are priorities and train-health fixed, irrespective of the context? What values constitute acceptable delays to a
low-priority train? Generating good train-schedules will require a combination of quantitative knowledge of
a train's running times and qualitative knowledge about the train in isolation, and in relation to other trains.
In this paper, we propose a heuristic search method, that comes under the broad category of an
estimation distribution algorithm (EDA). EDAs iteratively generates better solutions to the optimisation problem
using machine-constructed models. Usually EDA's have used generative probabilistic models, such as Bayesian
Networks, where domain-knowledge 
needs to be translated into prior distributions and/or network topology. In this paper,
we are concerned with problems for such a translation is not evident. Our interest in ILP is that it presents
perhaps one the most flexible ways to use domain-knowledge when constructing models.

What can this form of optimisation do differently? First, there is the straightforward difference
to standard optimisation, arising from the use of domain-knowledge in first-order logic. Traditionally,
optimisation methods have required domain knowledge to be in the form of linear inequalities. This quickly becomes
complicated. For example,  $y = x_1 \oplus x_2$ requires the
inequalities $y \leq x_1 + x_2 $ $\wedge$ $y \leq 2 - x_1 - x_2$ $\wedge$
$y \geq x_1 - x_2$ $\wedge$ $y \geq x_2 - x_1$ $\wedge$
$y \geq 0$ $\wedge$ $y \leq 1$. As a statement in logic, the relation
is clearly trivial: so, we would expect to do better on problems
for which domain-knowledge is far easier to express in logical form than as linear constraints (of course, one
could consider non-linear constraints, but then the
optimisation problem becomes much harder). Secondly, there is the difference arising from 
constructing models in first-order logic. Most probabilistic models used in EDA only allow
use models that involve statements about propositions.
This restricts the expressivity of the models, or requires large numbers of propositions representing
pre-defined relations. We would therefore expect to to do better on problems that require models
that involve relationships amongst background predicates that are not easy to know beforehand.

The rest of the paper is organised as follows. Section \ref{sec:eda-for-optimization} provides a brief description
of the EDA method we use for optimisation problems. Section \ref{sec:ilpeda} describes how ILP can be used within the iterative loop
of an EDA, including a procedure for sampling data instances entailed by the ILP-theory. Section \ref{sec:expt} describes an empirical evaluation 
followed by conclusions in Section \ref{sec:concl}.

\section{EDA for Optimisation}
\label{sec:eda-for-optimization}

The basic EDA approach we use is the one proposed by the MIMIC algorithm \cite{mimic}. Assuming
that we are looking to minimise an objective function $F(\textbf{x})$, where $\textbf{x}$ is
an instance from some instance-space ${\cal X}$, the approach first constructs an appropriate
machine-learning model to discriminate between samples of lower and higher value, i.e., $F(\mathbf{x}) \leq \theta$ and
$F(\mathbf{x}) > \theta$, and then sampling from this model to generate a population for the next iteration,
while also lowering $\theta$. This is described by the procedure in Fig.~\ref{fig:eoms}.

\begin{figure}[htb]
\begin{description}
\item[Procedure EOMS:] Evolutionary Optimisation using Model-Assisted Sampling
\begin{enumerate}
    \item Initialize population $P:=\{\textbf{x}_i\}$; $\theta:=\theta_0$
    \item while not converged do
        \begin{enumerate}
            \item for all $\mathbf{x}_i$ in $P$ $label(\textbf{x}_i)$ := 1 if
            $F(\textbf{x}_i) \leq \theta$ else $label(\textbf{x}_i)$ := 0
            \item train model $M$ to discriminate between 1 and 0 labels
                    i.e.,
                    $P(\textbf{x}: label(\textbf{x}) = 1 | M ) > P(\textbf{x}: label(\textbf{x}) = 0 | M )$
            \item regenerate $P$ by repeated sampling using model $M$
            \item reduce threshold $\theta$
        \end{enumerate}
\item return $P$
\end{enumerate}
\end{description}

\caption{Evolutionary optimisation using machine-learning models to guide sampling.}
\label{fig:eoms}
\end{figure}


\subsection{ILP-assisted EDA for Optimisation}
\label{sec:ilpeda}

We propose the use ILP as the model construction technique in the EOMS procedure,
since it provides an extremely flexible way to
construct models using domain-knowledge. On the face of it, this would seem to pose a
difficulty for the sampling step: how are we to generate new instances that are entailed by an
ILP-constructed model? There are two straightforward options. First, if we have
an enumerator of the instance space ${\cal X}$, then we could resort to a form
of rejection-sampling. Second,
we can restrict ILP-theories for any predicate to generative clauses, which allows
the theories to be used generatively.\footnote{A syntactic
way to do this is by adding constraints to the body of the clause that
impose range (that is, type) restrictions on the variables in the head:
see \cite{muggfeng:golem}.}, using standard logic-programming inference machinery to generate instances of
the success-set of each predicate. Instances obtained in this manner are selected with some probability to
achieve a non-uniform sampling of the success-set. Both these methods are
viable for the purposes of this paper, but in general, we expect that more sophisticated ways of sampling would be needed: see for
example \cite{cussens:slp}. The procedure EOIS in Fig.~\ref{fig:eois} is a refinement of the
EOMS procedure above.

\begin{figure}[htb]
\begin{description}
    \item[Procedure EOIS:] Evolutionary Optimisation using ILP-Assisted Sampling
    \vspace*{0.1in}
    \item[Given:] 
        (a) Background knowledge $B$;
        (b) an upper-bound $\theta^*$ on the cost of acceptable solutions;
        (c) a decreasing sequence of cost-values $\theta_1, \theta_2, \ldots, \theta_n$ s.t.
        $\theta_1 \geq \theta^* \geq \theta_n$; and
        (d) an upper-bound on the sample size $n$
\vspace*{0.1in}
\begin{enumerate}
    \item Let $M_0$:= $\emptyset$ and $P_0$:= $sample(n,M_0,B)$
    \item Let $k=1$ 
    \item while ($\theta_k \geq \theta^*$) do
        \begin{enumerate}
            \item ${E_k}^+$ :=
            $\{\textbf{x}_i: \textbf{x}_i \in P ~\mathrm{and}~F(\textbf{x}_i) \leq \theta_k\}$ and
            ${E_k}^-$ := $\{\textbf{x}_i: \textbf{x}_i \in P ~\mathrm{and}~F(\textbf{x}_i) > \theta_k\}$
            \item $M_k$ := $ilp(B,{E_k}^+,{E_k}^-)$
            \item $P_k$:= $sample(n,M_k,B)$
            \item increment $k$
        \end{enumerate}
    \item return $P_{k-1}$
\end{enumerate}
\end{description}
\caption{Evolutionary optimisation using ILP models to guide sampling.}
\label{fig:eois}
\end{figure}

In EOIS, $ilp(B,E^+,E^-)$ is an ILP algorithm that returns a theory
$M$ s.t. $B \wedge M \models E^+$;
$B \wedge M$ is inconsistent with the $E^-$ only to the extent allowed by
constraints in $B$; and $sample(n,M,B)$ returns a set of at most $n$ instances entailed
by $B \wedge M$, if $M \neq \emptyset$. If $M = \emptyset$, it returns a random selection
of $n$ instances from the instance-space. In general, we require $sample$ to be draw instances
from the success-set of the ILP-constructed theory, since these are the ``good'' solutions
entailed by the model on each iteration (see \cite{cussens:slp} for techniques for doing
this). Here, we make do by
providing an initial sample
to EOIS as input when $M = \emptyset$ (to prevent biasing future iterations:
this sample is obtained by uniform random selection
from the instance space). For subsequent steps, we assume the availability
of a generator that returns a sample of
the success-set of target-predicate. That is,
when $M \neq \emptyset$, $sample$ returns the set $S$ =
$\{e_i: 1 \leq i \leq n ~\mathrm e_i \in {\cal X}~\mathrm{and}~B \wedge M \vdash e ~\mathrm{and}~Pr(e_i) \geq \delta\}$,
where $\delta$ is some probability threshold (correctly therefore $sample(n,M,B)$
should be $sample(n,\delta,M,B)$). Thus if $\delta = 1$, the first
$n$ instances derived (or fewer if there are less) using SLD-resolution with
$B \wedge M$ will be selected.
Finally, we note that on iterations $k \geq 1$, we can use
the data $P_0 \cup \cdots \cup P_{k-1}$ to obtain training examples ${E_k}^+$ and ${E_k}^-$
since the actual costs for the $P$'s have
have already been computed. For clarity, this detail has been omitted.

\section{Empirical Evaluation}
\label{sec:expt}

\subsection{Aims}
\label{sec:aim}

Our aims in the empirical evaluation are to investigate the
following conjectures:
\begin{enumerate}
\item[(1)] On each iteration, the EOIS procedure will yield
        better samples than simple random sampling of the instance-space; and
\item[(2)] On termination, the EOIS procedure will yield more
        near-optimal instances than simple random sampling of
        the same number of instances as used for constructing
        the model.
\end{enumerate}

It is relevant here to clarify what the comparisons are intended in the statements above. Conjecture (1) is essentially a statement
about the gain in precision obtained by using the model. Let us denote
$Pr(F(\textbf{x}) \leq \theta)$ the probability of generating an instance $\mathbf{x}$ with cost at
most $\theta$ without a model to guide sampling (that is, using simple random
sampling of the instance space), and
by $Pr(F(\textbf{x}) \leq \theta|M_{k,B})$ the probability of obtaining such an instance with
an ILP-constructed model $M_{k,B}$ obtained on iteration $k$ of the EOIS
procedure using some domain-knowledge $B$.
(note if $M_{k,B} = \emptyset$, then we will mean
$Pr(F(\textbf{x}) \leq \theta|M_{k,B})$ = $Pr(F(\textbf{x}) \leq \theta)$).
Then for (1) to hold, we would require
$Pr(F(\textbf{x}) \leq \theta_k | M_{k,B}) > Pr(F(\textbf{x}) \leq \theta_k)$.
given some relevant $B$. We will estimate the probability on the lhs from
the sample generated using the model, and the probability on the rhs from
the datasets provided.

Conjecture (2) is related to the gain in recall obtained by using the model,
although it is more practical to examine actual numbers of near-optimal instances
(true-positives in the usual terminology). We will compare the numbers of near-optimal
in the sample generated by the model to those obtained using random sampling.


\subsection{Materials}
\label{sec:materials}

\subsubsection{Data}
We use two synthetic datasets, one arising from the KRK chess endgame (an endgame with just White King, White Rook and Black King
on the board), and the other a restricted, but nevertheless hard $5 \times 5$ job-shop scheduling (scheduling 5 jobs taking varying lengths of
time onto 5 machines, each capable of processing just one task at a time).

The optimisation problem we examine for the KRK endgame is to predict the depth-of-win with optimal play \cite{bain:krkwin}.
Although aspect of the endgame has not been as popular in ILP as task of predicting ``White-to-move position is illegal''
\cite{bain:krkillegal,compr:ilp92}, it 
offers a number of advantages as a {\em Drosophila\/} for optimisation problems of the kind we are interested.
First, as with other chess endgames, KRK-win is a complex, enumerable domain for which
there is complete, noise-free data. Second, optimal ``costs'' are known for all data instances.
Third, the problem has been studied by
chess-experts at least since Torres y Quevado built a machine, in 1910, capable of playing the KRK endgame. This
has resulted in a substantial amount of domain-specific knowledge. We direct the reader to \cite{breda:thesis}
for the history of automated methods for the KRK-endgame. For us, it suffices to treat the problem as a form
of optimisation, with the cost being the depth-of-win with Black-to-move, assuming minimax-optimal play.
In principle, there are ${64}^3 ~\approx 260,000$ possible positions for the KRK endgame,
not all legal. Removing illegal
positions, and redundancies arising from symmetries of the board reduces the size of the
instance space to about $28,000$ and the distribution shown in
Fig.~\ref{fig:distr}(a). The sampling task here is to generate instances with depth-of-win equal to $0$.
Simple random sampling has a probability of about $1/1000$ of generating such an instance once redundancies
are removed.

The job-shop scheduling problem is less controlled than the chess endgame, but is nevertheless representative
of many real-life applications (like scheduling trains), which are, in general, known to be computationally
hard. We use a job-shop problem with five jobs,
each consisting of five tasks that need to be executed in order. These 25 tasks are to be 
performed using 5 machines, each capable of performing a particular task, albeit for any of the jobs.
A $5 \times 5$ matrix defines how long task $j$ of job $i$ takes to execute on machine $j$.

\begin{figure}
\begin{minipage}[h]{0.39\textwidth}
\centering
{\scriptsize{
\begin{tabular}{|cl|cl|} \hline
Cost & Instances   & Cost & Instances \\ \hline
0     & 27 (0.001) & 9     & 1712 (0.196) \\
1     & 78 (0.004)  & 10     &1985 (0.267) \\
2     & 246 (0.012) & 11    & 2854 (0.368) \\
3     & 81  (0.152) & 12    & 3597 (0.497) \\
4     & 198 (0.022) & 13    & 4194 (0.646) \\
5     & 471 (0.039) & 14    & 4553 (0.808)\\
6     & 592 (0.060) & 15    &  2166 (0.886) \\
7     & 683 (0.084) & 16    & 390 (0.899)  \\
8     & 1433 (0.136)   & draw  & 2796 (1.0) \\ \hline 
\multicolumn{4}{l}{Total Instances: 28056} \\
\end{tabular}
}}
\begin{center}
(a) Chess
\end{center}
\end{minipage}
\begin{minipage}[h]{0.60\textwidth}
\vspace*{1cm}
\centering
{\scriptsize{
\begin{tabular}{|cl|cl|} \hline
Cost & Instances & Cost & Instances \\ \hline
400--500 & 10 (0.0001) & 1000--1100   & 24067 (0.748) \\
500--600 & 294 (0.003) & 1100--1200  &  15913 (0.907) \\
600--700 & 2186 (0.025)  & 1200--1300  & 7025 (0.978) \\
700--800 & 7744 (0.102) & 1300--1400  &  1818 (0.996) \\
800--900 & 16398 (0.266) & 1400--1500 & 345 (0.999) \\
900--1000& 24135 (0.508) & 1500--1700 & 66  (1.0) \\ \hline 
\multicolumn{4}{l}{Total Instances: 100000} \\
\end{tabular}
}}
\begin{center}
(b) Job-Shop
\end{center}
\end{minipage}
\caption{Distribution of cost values. Numbers in parentheses are
    cumulative frequencies.}
\label{fig:distr}
\end{figure}

\noindent

Data instances for Chess are in the form of 6-tuples, representing the rank and file (X and Y values) of the 3 pieces
involved. At each iteration $k$ of the EOIS procedure, some instances with depth-of-win $\leq \theta_k$ and the rest
with depth-of-win $> \theta_k$ are used to construct a model.\footnote{The $\theta_k$ values are pre-computed
assuming optimum play. We note that when constructing a model on iteration $k$, it is permissible to use
all instances used on iterations $1,2,\ldots,(k-1)$ to obtain data for model-construction.}

Data instances for Job-Shop are in the form of schedules defining the sequence in which tasks of different jobs are
performed on each machine, along with the total cost (i.e., time duration) implied by the schedule. 
On iteration $i$ of the EOIS procedure,
models are to be constructed to predict if the cost of schedule will be $\leq \theta_i$ or otherwise.\footnote{The
total cost of a schedule includes any idle-time, since for each job, a task
before the next one can be started for that job. Again, on iteration $i$,
it is permissible to use data from previous iterations.}

\subsubsection{Background Knowledge}

For Chess, background predicates encode the following (WK denotes, WR the White Rook, and BK the Black King):
(a) Distance between pieces WK-BK, WK-BK, WK-WR;
(b) File and distance patterns: WR-BK, WK-WR, WK-BK;
(c) ``Alignment distance'': WR-BK;
(d) Adjacency patterns: WK-WR, WK-BK, WR-BK;
(e) ``Between'' patterns: WR between WK and BK, WK between WR and BK, BK
    between WK and WR;
(f) Distance to closest edge: BK;
(g) Distance to closest corner: BK;
(h) Distance to centre: WK; and
(i) Inter-piece patterns: Kings in opposition, Kings almost-in-opposition,
    L-shaped pattern.
We direct the reader to \cite{breda:thesis} for the history of using these concepts, and their definitions.

For Job-Shop, background predicates encode:
(a) schedule job $J$ ``early'' on machine $M$ (early means first or second);
(b) schedule job $J$ ``late'' on machine $M$ (late means last or second-last);
(c) job $J$ has the fastest task for machine $M$;
(d) job $J$ has the slowest task for machine $M$;
(e) job $J$ has a fast task for machine $M$ (fast means the fastest or second-fastest);
(f) Job $J$ has a slow task for machine $M$ (slow means slowest or second-slowest);
(g) Waiting time for machine $M$;
(h) Total waiting time;
(i) Time taken before executing a task on a machine. Correctly, the predicates for (g)--(i)
encode upper and lower bounds on times, using the standard inequality predicates $\leq$ and $\geq$.

\subsubsection{Algorithms and Machines}

The ILP-engine we use is Aleph (Version 6, available from A.S. on request). 
All ILP theories were constructed on an Intel Core i7 laptop computer, using
VMware virtual machine running Fedora 13, with an allocation of 2GB for
the virtual machine. The Prolog compiler used was Yap,
version 6.1.3\footnote{\urltilda{http://www.dcc.fc.up.pt/~vsc/Yap/}}.
\subsection{Method}
\label{sec:method}

Our method is straightforward:

\begin{itemize}
\item[] For each optimisation problem, and domain-knowledge $B$:
    \begin{itemize}
        \item[] Using a sequence of threshold values
                $\langle \theta_1, \theta_2, \ldots, \theta_n \rangle$
        on iteration $k$ ($1 \leq k \leq n$) for the EOIS procedure:
        \begin{enumerate}
            \item Obtain an estimate of $Pr(F(\textbf{x}) \leq \theta_k)$ using
                                a simple random sample from the instance space;
            \item Obtain an estimate of $Pr(F(\textbf{x}) \leq \theta_k|M_{k,B})$
                                    by constructing an ILP model for
                                    discriminating between $F(\textbf{x}) \leq \theta_k$ and
                                    $F(\textbf{x}) > \theta_k$
            \item Compute the ratio of
                                $Pr(F(\textbf{x}) \leq \theta_k|M_{k,B})$
                                to $P(F(\textbf{x}) \leq \theta_k)$
         \end{enumerate}
    \end{itemize}
\end{itemize}

\noindent
The following details are relevant:
\begin{itemize}
\item The sequence of thresholds for Chess are $\langle 8, 4, 0 \rangle$. For
    Job-Shop, this sequence is $\langle 1000, 750, 600 \rangle$; Thus,
    $\theta^*$ = 0 for Chess and $600$ for Job-Shop, which means we
    require exactly optimal solutions for Chess.
\item Experience with the use of ILP engine used here (Aleph)
    suggests that the most sensitive parameter is the one defining a lower-bound on the 
        precision of acceptable clauses (the $minacc$ setting in Aleph).
        We report experimental results obtained with $minacc=0.7$, which has
        been used in previous experiments with the KRK dataset.
        The background knowledge for Job-Shop does not appear to
        be sufficiently powerful to allow the identification of
        good theories with short clauses. That is, the usual Aleph
        setting of upto 4 literals per clause leaves most of the
        training data ungeneralised. We therefore allow an
        upper-bound of upto 10 literals for Job-Shop, with
        a corresponding increase in the number of search nodes to 10000
        (Chess uses the default setting of 4 and 5000 for these parameters).
\item In the EOIS procedure, the bound on sample size $n$ is 1000.
        The initial sample is obtained using a uniform
    distribution over all instances. Let us call this
        $P_0$. On the first iteration of EOIS ($k=1$), the datasets
        ${E_1}^+$ and ${E_1}^-$ are obtained by computing the (actual) costs for
        instances in $P_0$, and an ILP model $M_{1,B}$, or simply $M_1$, constructed.
        To obtain a sample of instances entailed by the model $M_{k,B}$
        we use the $sample$ function with a $\delta$ value of $1.0$.
        That is, the first $n$ unique instances (or fewer, if less) obtained
        by employing SLD-resolution on $B \wedge M_{k,B}$ are taken as
        the sample $P_k$. For Chess, it has been possible to ensure
        that the logic-programs involved are generative. Thus, we
        are able to use $M_{k,B}$ directly as a generator of
        instances entailed by the $B \wedge M_{k,B}$. For Job-Shop,
        we employ rejection-sampling instead. That is, we randomly
        draw from the instance-space, and then check to see if it
        is entailed by $B \wedge M_{k,B}$. Both approaches have
        not proved to be especially inefficient, probably because the
        instance-spaces are small.
        On each iteration $k$, an estimate of $Pr(F(\textbf{x}) \leq \theta_k)$ can be
        obtained from the empirical frequency distribution of instances with values
        $\leq \theta_k$ and $> \theta_k$. For the synthetic problems here,
        these estimates are in Fig.~\ref{fig:distr}.
        For $Pr(F(\textbf{x}) \leq \theta_k|M_{k,B})$, we use
        obtain the frequency of $F(\textbf{x}) \leq \theta_k$ in $P_k$
\item Readers will recognise that the ratio of
        $Pr(F(\textbf{x}) \leq \theta_k|M_{k,B})$
        to $P(F(\textbf{x}) \leq \theta_k)$
    is equivalent to computing the gain in precision obtained
    by using an ILP model over random selection. Specifically, if this ratio
    is approximately $1$, then there is no value in using the ILP model. The
    probabilities computed also provide one way of estimating sampling efficiency
    of the models (the higher the probability, the fewer samples will be needed
    to obtain an instance $\mathbf{x}$ with $F(\mathbf{x}) \leq \theta_k$).
\end{itemize}

\subsection{Results}
\label{sec:results}

Results relevant to conjectures (1) and (2) are tabulated in
Fig.~\ref{fig:results} and Fig.~\ref{fig:results2}.
The principal conclusions that can drawn from the results are these:
\begin{enumerate}
\item[(1)] For both problems, and every threshold value $\theta_k$,
    the probabilty of obtaining instances
        with cost at most $\theta_k$ with model-guided samoling is
        substantially higher than without a model. This provides evidence
        that model-guided sampling results in better samples than
        simple random sampling (Conjecture 1);
\item[(2)] For both problems and every threshold value $\theta_k$, 
        samples obtained with model-guided sampling contain a 
        substantially number of near-optimal instances
        than samples obtained with a model (Conjecture 2)
\end{enumerate}

\noindent
We note also that all results have been obtained by sampling a small
portion of the instance space (about 10 \% for Chess, and about 3 \% for Job-Shop).
        
\begin{figure}
\begin{minipage}[h]{0.49\textwidth}
\centering
\begin{tabular}{|l|c|c|c|}
\hline
 Model& \multicolumn{3}{|c|}{$Pr(F(\mathbf{x}) \leq \theta_k|M_k)$} \\ \cline{2-4}
      & $k=1$ & $k=2$ & $k=3$ \\ \hline
None  & $0.136$ & $0.022$ & $0.001$ \\[6pt]
ILP   & $0.816$ & $0.462$ & $0.409$ \\
      & ($6.0$)   & ($21.0$)  & ($409.0$) \\ \hline 
\end{tabular}
\begin{center}
(a) Chess
\end{center}
\end{minipage}
\begin{minipage}[h]{0.5\textwidth}
\centering
\begin{tabular}{|l|c|c|c|}
\hline
 Model & \multicolumn{3}{|c|}{$Pr(F(\mathbf{x}) \leq \theta_k|M_k)$} \\ \cline{2-4}
       & $k=1$ & $k=2$ & $k=3$ \\ \hline
None  & $0.507$ & $0.025$ & $0.003$ \\[6pt]
ILP   & $0.647$ & $0.171$ & $0.080$ \\
        & ($1.3$)   & ($6.8$)  & ($26.7$) \\ \hline 
\end{tabular}
\begin{center}
(b) Job-Shop
\end{center}
\end{minipage}
\caption{Probabilities of obtaining good instances $\mathbf{x}$ for each iteration
    $k$ of the EOIS procedure. That is, the column $k=1$ denotes
    $P(F(\mathbf{x}) \leq \theta_1$ after iteration $1$; the column $k=2$ denotes
    $P(F(\mathbf{x}) \leq \theta_2$ after iteration $2$ and so on.
        In effect, this is an estimate of the precision when predicting
        $F(\mathbf{x}) \leq \theta_k$.
    ``None'' in the model column stands for
    probabilities of the instances, corresponding to simple random sampling
        ($M_k = \emptyset$).
    The number in parentheses below each ILP entry denotes the
        ratio of that entry against
        the corresponding entry for ``None''. This represents the gain in
        precision of using the ILP model over simple random sampling.}
\label{fig:results}
\end{figure}

\begin{figure}
\begin{minipage}[h]{0.49\textwidth}
\centering
\begin{tabular}{|l|c|c|c|} \hline
Model & \multicolumn{3}{|c|}{Near-Optimal Instances} \\ \cline{2-4}
      & $k=1$ & $k=2$ & $k=3$ \\ \hline
None  & 1/27   & 2/27 & 3/27 \\[6pt]
ILP   &  11/27 &  22/27 & 27/27 \\
      & ($1000$) & ($1964$) & ($2549$) \\ \hline
\end{tabular}
\begin{center}
(a) Chess
\end{center}
\end{minipage}
\begin{minipage}[h]{0.5\textwidth}
\centering
\begin{tabular}{|l|c|c|c|} \hline
Model & \multicolumn{3}{|c|}{Near-Optimal Instances} \\ \cline{2-4}
      & $k=1$ & $k=2$ & $k=3$ \\ \hline
None  & 3/304 & 6/304 & 9/304 \\[6pt]
ILP   & 6/304 & 28/304 & 36/304 \\
       & ($1000$) & ($1987$) & ($2895$) \\ \hline
\end{tabular}
\begin{center}
(b) Job-Shop
\end{center}
\end{minipage}
\caption{Fraction of near-optimal instances ($F(\textbf{x}) \leq \theta^*)$
    generated on each iteration of EOIS.  In effect, this is
    an estimate of the recall (true-positive rate, or sensitivity)
    when predicting $F(\textbf{x}) \leq \theta^*$. The fraction $a/b$
    denotes that $a$ instances of $b$ are generated. The numbers
    in parentheses denote the number of training instances used by the ILP engine.
    The values with ``None'' are the numbers expected
        by sampling the same number of training instances  used for training the
        ILP engine.}
\label{fig:results2}
\end{figure}

\noindent
We now examine the result in more detail. It is evident that the performance on
the Job-Shop domain is not as good as on Chess. The natural question that arises is: Why is this so?
We conjecture that this is a consequence of the background knowledge for Job-Shop not
being as relevant to low cost values, as was the case for Chess. Some evidence for this
was already apparent when we had to increase the lengths of clauses allowed for the ILP engine
(this is usually a sign that the background knowledge is somewhat low-level). In contrast,
with Chess, some of the
concepts refer specifically to ``cornering'' the Black King, with a view of ending the game
as soon as possible. We would expect these predicates to be especially useful for positions
at depths-of-win near 0. Evidence of the unreliable performance of the EOIS procedure in Chess, with
irrelevant background knowledge is in Fig.~\ref{fig:chessb0results}. These results suggest
a refinement to the conclusions we can draw from the use of
EOIS, namely: we expect the EOIS procedure to be less effective if the background
knowledge is not very relevant to low-cost solutions.

Finally, we note that the experiments with synthetic data have ignored
an important aspect of the optimisation problem, namely the time taken to obtain the value of
the objective function for each data instance. Clearly, there is a trade-off between the time taken to
construct a model, and the time taken to simply draw instances without a model.
To address this trade-off we would have estimate the number of instances that need
to be randomly sampled to obtain the same numbers of near-optimal
instances as with model-assisted EDA; and to compare: (a) the time to obtain
values of the objective function for randomly-sampled instances; and
(b) the time taken to obtain the corresponding values for the training
data used to construct models and the total time taken for
model construction. For model-construction to be beneficial,
clearly the time in (b) has to be less than (a).
For the problems here, random sampling we require approximately 4 (JobShop) to 10 (Chess)
times as many samples to obtain the same numbers of near-optimal instances.
The times for theory-construction are small enough to expect
that (b) is less than (a) for these ratios.

\section{Concluding Remarks}
\label{sec:concl}

It is uncommon to see ILP applied to optimisation problems. 
The use of ILP-constructed theories within an evolutionary optimisation
procedure is one answer to the question of how to use ILP for optimisation (but 
not the only answer: recent work in \cite{ohwada:optim}, for example, suggests using
ILP-constructed clauses as soft-constraints to a constraint solver). This requires the
ILP model to be used generatively, which is not common practice in ILP. In this paper we have
been able to do this by a combination of careful definition of the background knowledge,
and adding range-restrictions to clauses constructed by ILP. Our results provide evidence
that this combination of ILP and logic-programming provides one way of incorporating complex,
but relevant domain-knowledge into evolutionary optimisation.

Concerning the specific problems examined here, it is possible
that we could have directly constructed a model discriminating near-optimal instances
from the rest using ILP alone. The focus of this paper however is on a different
question, namely, whether evolutionary optimisation 
methods can benefit from the use of ILP. The results here should therefore
be seen as evidence of improvements possible in an EDA technique when it
includes ILP-assisted models. In turn, this evidence
could be of relevance for problems where ILP models
alone would be insufficient, and we would have to
resort to sampling-based methods.

There are several ways in which the work here could be extended.
The most immediate is to examine ways of sampling by using techniques
developed in probabilistic ILP. Indeed, 
the principal conjecture of the paper is that the use of
models constructed by any form of learning that allows the
inclusion domain-knowledge can greatly improve
the sampling efficiency of EDA methods. We have provided evidence
for this conjecture using a classical ILP method. Given these
results, we can expect first-order learning that is capable
of using domain-knowledge and constructing rules that can
allow a non-uniform sampling of ground instances (for example, through
the incorporation of probabilities with the rules) will provide
even better. We would recommend this as the next step in this line of work.

It is of interest also to consider whether there are any gains
to be made by re-use of theories (currently, we re-use data, but re-learn
theories from scratch on each new iteration of the EOIS procedure). There
is the straightforward approach to this, of simply providing theories
constructed earlier as background knowledge for subsequent iterations. A more
ambitious variant would retain some portions of the earlier theory (those
clauses that entail the current set of positive instances and and none, or only few, of the negative instances,
for example), thus reducing the model-construction effort.

\begin{figure}
\begin{minipage}[h]{0.49\textwidth}
\vspace*{1cm}
\centering
\begin{tabular}{|l|c|c|c|}
\hline
 Background & \multicolumn{3}{|c|}{$Pr(F(\mathbf{x}) \leq \theta_k|M_{k,B})$} \\ \cline{2-4}
 $B$        & $k=1$ & $k=2$ & $k=3$ \\ \hline
$B_{low}$    & $0.658$ & $0.417$ & $0.0$ \\[6pt]
$B_{high}$   & $0.816$ & $0.462$ & $0.409$ \\ \hline
\end{tabular}
\begin{center}
(a)
\end{center}
\end{minipage}
\begin{minipage}[h]{0.5\textwidth}
\centering
\begin{tabular}{|l|c|c|c|} \hline
Background & \multicolumn{3}{|c|}{Near-Optimal Instances} \\ \cline{2-4}
$B$        & $k=1$ & $k=2$ & $k=3$ \\ \hline
$B_{low}$  &  17/27 & 4/27 & 0/27 \\
      & ($1000$) & ($1959$) & ($2581$) \\[6pt]
$B_{high}$  & 11/27   & 17/27 & 27/27 \\
            & ($1000$) & ($1964$) & ($2549$) \\ \hline
\end{tabular}
\begin{center}
(b)
\end{center}
\end{minipage}
\caption{Precision (a), and recall of near-optimal instances (b) of the EOIS
    procedure with background knowledge of low relevance to near-optimal
    solutions. The results are for Chess, with $B_{low}$ denoting background
    predicates that simply define the geometry of the board, using the predicates
    $less\_than$ and $adjacent$. These predicates form the background knowledge
    for most ILP applications to the problem of detecting illegal positions in
    the KRK endgame. $B_{high}$ denotes background used previously, with
                high relevance to low depths-of-win. The numbers in parentheses in (b)
                are the number of training instances as before.} 
\label{fig:chessb0results}
\end{figure}

\subsection*{Acknowledgments}

A.S. is a Visiting Professor at the Department of Computer Science,
University of Oxford; and Visiting Professorial Fellow at the School
of CSE, UNSW.

\bibliographystyle{plain}
\bibliography{trefs}

\end{document}